\documentclass[10pt,twocolumn,letterpaper]{article}

\usepackage{cvpr}
\usepackage{times}
\usepackage{epsfig}
\usepackage{graphicx}
\usepackage{amsmath}
\usepackage{amssymb}
\usepackage{leftidx}

\usepackage{multirow}
\usepackage{multicol}
\usepackage{array}
\usepackage{textcomp}
\usepackage{color}
\usepackage{placeins}
\usepackage{caption}

\usepackage{lipsum}
\usepackage[breaklinks=true,bookmarks=false]{hyperref}

\cvprfinalcopy % *** Uncomment this line for the final submission

\def\cvprPaperID{1810} % *** Enter the CVPR Paper ID here

\ifcvprfinal\fi
\begin{document}

%%%%%%%%% TITLE
\title{Layout-Graph Reasoning for Fashion Landmark Detection}

\author{Weijiang Yu$^{1}$,~~ Xiaodan Liang$^{1,2*}$,~~ Ke Gong$^{1,2}$,~~ Chenhan Jiang$^{1}$,\\ Nong Xiao$^1$,~~ Liang Lin$^{1,2}$\\
$\leftidx^1$Sun Yat-sen University,~~ $\leftidx^2$DarkMatter AI Research\\
{\tt\small weijiangyu8@gmail.com, \tt\small xdliang328@gmail.com, \tt\small kegong936@gmail.com,}\\
	{\tt\small jchcyan@gmail.com, \tt\small xiaon6@sysu.edu.cn, \tt\small linliang@ieee.org}
}
\maketitle
\newcommand\blfootnote[1]{% 
	\begingroup 
	\renewcommand\thefootnote{}\footnote{#1}% 
	\addtocounter{footnote}{-1}% 
	\endgroup 
}
\thispagestyle{empty}
%\pagestyle{empty}

%%%%%%%%% ABSTRACT
\begin{abstract}
   Detecting dense landmarks for diverse clothes, as a fundamental technique for clothes analysis, has attracted increasing research attention due to its huge application potential. However, due to the lack of modeling underlying semantic layout constraints among landmarks, prior works often detect ambiguous and structure-inconsistent landmarks of multiple overlapped clothes in one person. In this paper, we propose to seamlessly enforce structural layout relationships among landmarks on the intermediate representations via multiple stacked layout-graph reasoning layers. We define the layout-graph as a hierarchical structure including a root node, body-part nodes (e.g. upper body, lower body), coarse clothes-part nodes (e.g. collar, sleeve) and leaf landmark nodes (e.g. left-collar, right-collar). Each Layout-Graph Reasoning(LGR) layer aims to map feature representations into structural graph nodes via a Map-to-Node module, performs reasoning over structural graph nodes to achieve global layout coherency via a layout-graph reasoning module, and then maps graph nodes back to enhance feature representations via a Node-to-Map module. The layout-graph reasoning module integrates a graph clustering operation to generate representations of intermediate nodes (bottom-up inference) and then a graph deconvolution operation (top-down inference) over the whole graph. Extensive experiments on two public fashion landmark datasets demonstrate the superiority of our model. Furthermore, to advance the fine-grained fashion landmark research for supporting more comprehensive clothes generation and attribute recognition, we contribute the first Fine-grained Fashion Landmark Dataset (FFLD) containing 200k images annotated with at most 32 key-points for 13 clothes types.
\end{abstract}
\let\thefootnote\relax\footnotetext{*Corresponding Author}
%\let\thefootnote\relax\footnotetext{This work was done when the first author does internship at sensetime}
%%%%%%%%% BODY TEXT
\begin{figure}[t]
	\centering\includegraphics[width=1.0\linewidth]{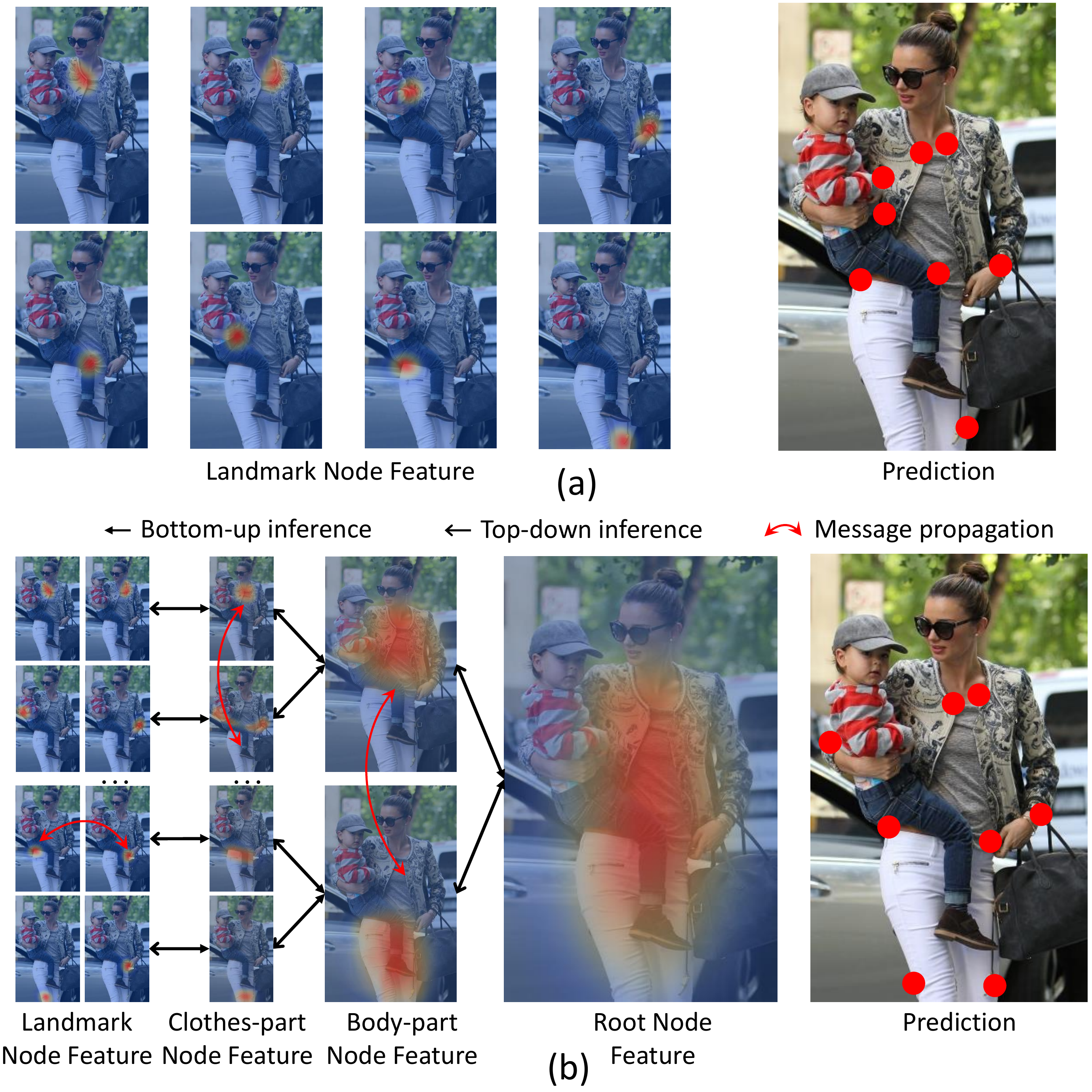}
	\vspace{-7mm}
	\caption{ 
		\textbf{(a)} Traditional DCNNs \cite{simonyan2014very} suffer from great performance drops when facing heavy overlapping of humans and clothes nesting because of lacking structural constraint and commonsense knowledge. (\eg the left waistline and right sleeve on the lady are wrongly predicted on the child.) \textbf{(b)} Our LGR provides graph-based inferences among landmarks, leveraging symmetric and hierarchical relations to constrain landmark's layout in one person.
	}
	\label{fig:result1}
	\vspace{-7mm}
\end{figure}
%%%%%%%%% BODY TEXT
\section{Introduction}
Fashion landmark detection that targets at localizing key-points at the functional regions of clothes (\eg collar, waistline), has attracted research attention and lots of demands driven by the boom of electronic commerce, such as clothing retrieval \cite{liu2016deepfashion,yamaguchi2015retrieving,di2013style,liu2012street}, clothes generation \cite{huang2015cross,WhereToBuyItICCV15} and fashion image classification \cite{liu2016deepfashion,wang2018attentive,han2017automatic}. To support these comprehensive high-level applications, the landmark detectors need to effectively deal with arbitrary clothing appearances, diverse clothes layouts and styles, multiply human pose, different lights and background clutters.

Recent research efforts on fashion landmark detection \cite{liu2015deep,trigeorgis2016mnemonic,liu2016deepfashion,yan2017unconstrained,liang2018look,yang2017learning,liu2016fashion} are mainly devoted to designing more advanced deep feature representations \cite{liu2015deep,liu2016deepfashion}, attention mechanism \cite{yan2017unconstrained}, pyramid module \cite{yang2017learning} and so on. These models are limited in regarding fashion landmark detection as an end-to-end regression problem and ignore rich semantic layout relations among different landmarks, such as symmetric relation (\eg left/right collar), subordinate relation (\eg left collar belongs to collar) and human commonsense (\eg one clothing generally owns a pair of sleeves). Consequently, unreasonable detection results that deviate from human and clothes layouts may be generated, as shown in Figure~\ref{fig:result1}(a).

Nevertheless, some researches have resorted to external guidance to enhance the interpretation of CNNs \cite{rothrock2011human,Chen_NIPS14,yang2016end,wang2018attentive,liang2018dynamic,lu2016visual}. For example, Yang \etal \cite{yang2016end} introduced to incorporate domain knowledge for explicitly facilitating features for better localization. Wang \etal \cite{wang2018attentive} proposed a grammar model in fashion visual understanding and used a bidirectional recurrent neural network for message passing. Fashion landmarks naturally lie in an underlying hierarchical structure which includes different levels of semantic nodes (e.g. body-part nodes, clothes-part nodes and leaf landmark nodes). However, they used a plain graph structure to represent knowledge and disregarded the intrinsic hierarchical and multi-level layouts of landmarks for better mining subordinate relations.

To address all above-mentioned issues, we propose to endow the deep networks with the capability of structural graph reasoning in order to make detected fashion landmarks be coherent with human and clothes layouts from a global perspective. We define a hierarchical layout-graph that encodes prior commonsense knowledge in terms of human body part layouts and clothes part layouts, consisting of a root node, body-part nodes (\eg upper body, lower body), coarse clothes-part nodes (\eg collar, sleeve) and leaf landmark nodes (\eg left-collar, right-collar). We then propose a novel Layout-Graph Reasoning (LGR) layer that is able to explicitly enforce hierarchical human-clothes layout constraints and semantic relations of fashion landmarks on deep representations for facilitating landmark detection. Our LGR layer is a general and flexible network layer which can be stacked and injected between any convolution layers, containing three modules: 1) a Map-to-Node module that maps convolutional features into each graph leaf node; 2) a layout-graph reasoning module to perform hierarchical graph reasoning over structural graph nodes to achieve global layout coherency;  3) a Node-to-Map module to learn appropriate associations between the evolved graph leaf nodes and convolutional features, which in turn enhance local feature representations with global reasoning.

Given graph node representations for leaf landmark nodes from the Map-to-Node module, our layout-graph reasoning module first performs a graph clustering operation to generate representations of intermediate nodes in the spirit of bottom-up inference, that is, propagating from (leaf landmark nodes)$\rightarrow$(clothes-part nodes)$\rightarrow$(body-part nodes)$\rightarrow$(root node). Then a graph deconvolution operation to evolve representations of bottom nodes guided by the higher-level structure nodes in the spirit of top-down inference, that is, (root node)$\rightarrow$(body-part nodes)$\rightarrow$(clothes-part nodes)$\rightarrow$(leaf landmark nodes). Benefiting from integrating graph clustering and graph deconvolution operations, our LGR layer enables to achieve global structural coherency and effectively enhance each landmark node representation for better predictions.

Moreover, existing fashion landmark datasets \cite{liu2016deepfashion,liu2016fashion,yan2017unconstrained} annotated with at most 8 landmarks for all clothes appearance. To advance the development of fine-grained domain knowledge in fashion landmark detection research, we contribute a new Fine-grained Fashion Landmark Dataset containing 200k images annotated with at most 32 key-points for 13 clothes types, named as FFLD. More details of FFLD are presented in supplementary material.

Our contributions are summarized in the following aspects: 

\textbf{1)} we propose a general Layout-Graph Reasoning (LGR) layer and incorporate multiply LGR layers into deep networks to seamlessly enforce structural layout relations among clothing landmarks on the intermediate representations to achieve global structure coherency. 

\textbf{2)} We define the layout-graph as a hierarchical structure for mining contextual graph semantic information from specific nodes to abstraction nodes. The graph clustering and graph deconvolution operation are integrated into each LGR layer for hierarchical graph reasoning. 

\textbf{3)} We construct the first Fine-grained Fashion Landmark Dataset (FFLD) that provides more comprehensive landmark annotations for diverse clothes types. 

\textbf{4)} Our model performs the superior ability compared with state-of-the-art approaches over two public fashion landmark datasets (\eg FLD \cite{liu2016fashion} and DeepFashion \cite{liu2016deepfashion}).

\section{Related Work}
\textbf{Fashion Landmark Detection and Localization.} 
Recently many research efforts have been devoted to joint localization and landmark detection \cite{liu2015deep,zhu2012face,Gong_2017_CVPR,trigeorgis2016mnemonic,wu2018look,yang2016end,Wei_2016_CVPR,yan2017unconstrained,liu2016deepfashion,liu2016fashion,wang2018attentive,newell2016stacked}. Newell \etal \cite{newell2016stacked} proposed a model for human pose estimation using a repeated pooling down and upsampling process to learn the spatial distribution of resolution. Liu \etal \cite{liu2016fashion} proposed deep fashion alignment using the pseudo-label scheme to enhance invariability of fashion landmark. Wang \etal \cite{wang2018attentive} captured kinematic and symmetry grammar of clothing landmark for mining geometric relationships among landmarks. They modeled grammar message passing processing as a bidirectional convolutional recurrent neural network for training in an end-to-end manner. 

The models of deep learning above demonstrate the powerful representations of neural networks. Few of them consider combining knowledge-guide information with fashion landmark detection in a hierarchical way. Motivated by Rothrock \etal \cite{rothrock2011human,kipf2017semi,ying2018hierarchical}, we build a hierarchical architecture to model global-local fashion landmark correlations for facilitating contextual information across each landmark.
\begin{figure*}
	\centering\includegraphics[width=1\linewidth]{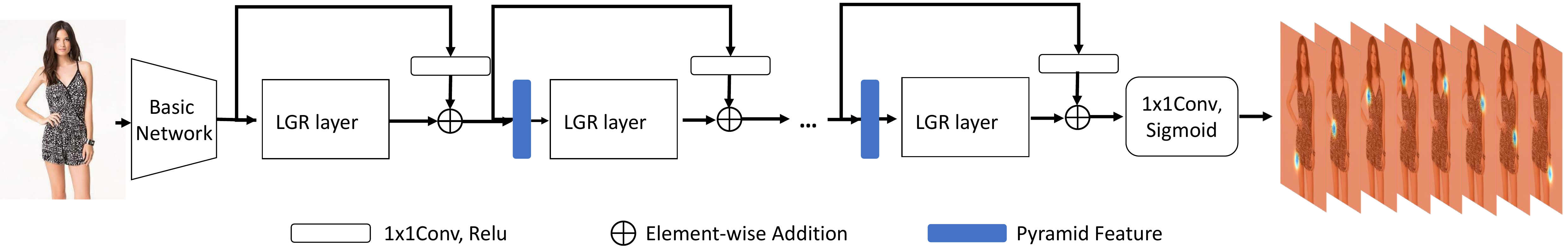}
	\vspace{-8mm}
	\caption{Illustration of our model that incorporates basic convolutional network for features extraction and stacked Layout-Graph Reasoning layers for structural graph reasoning. Residual addition processing and pyramid feature post-processing are appended between each stacked architecture for reducing bias and capturing rich representations across multiple scales. A 1$\times$ 1 convolution with $sigmoid$ activation function is utilized to produce final fashion landmark heatmaps. For better viewing of all figures in this paper, please see the original zoomed-in color pdf file.}
	\label{fig:framework}
	%	\vspace{-4mm}
\end{figure*}
\begin{figure*}[t]
	\centering
	\includegraphics[width=1.0\linewidth]{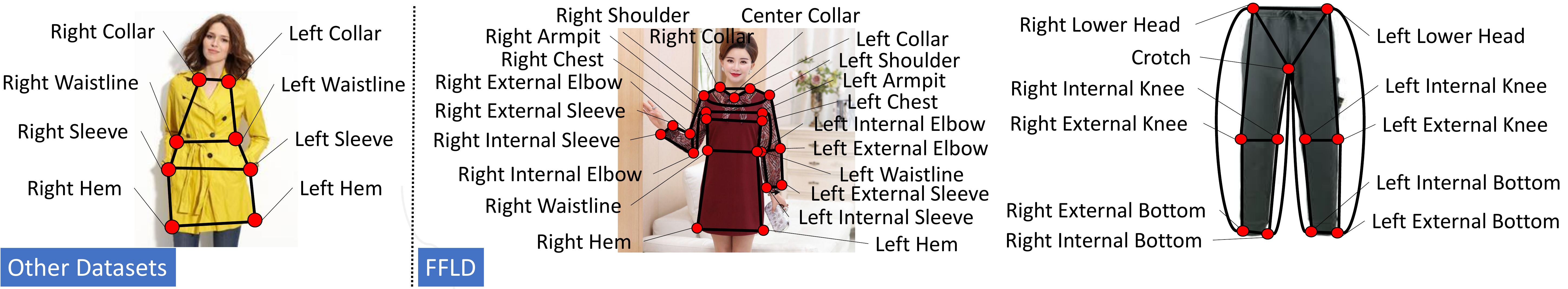}
	\vspace{-6mm}
	\caption{Layout-graph definitions for different fashion landmark datasets. Each leaf node (red circles) represents the position and type of one clothing landmark. Each leaf edge (black lines) indicates the correlations between landmark points.}
	\vspace{-4mm}
	\label{fig:graph_relation}
\end{figure*}

\textbf{Knowledge-guide Information in Graph.} 
Recently some research efforts model domain knowledge as graph for mining correlations among labels or objects in images, which has been proved effective in many tasks \cite{kipf2017semi,NIPS2016_6081,niepert2016learning,ying2018hierarchical,li2018factorizable,lu2016visual,rothrock2011human,battaglia2018relational,deng2014large,gan2016concepts}. Li \etal \cite{li2018factorizable} proposed a subgraph-based model for scene graph generation using bottom-up relationship inference of objects in images. Liang \etal \cite{liang2018dynamic} modeled semantic correlations using semantic neuron graph for explicitly incorporating semantic concept hierarchy during propagation. Yang \etal \cite{yang2016end} built a prior knowledge-guide graph for body part locations to well consider the global pose configurations. 

As far as we know, there is no work considering modeling layout-graph across all scales and levels in the layout among fashion landmarks. We incorporate cross-layer graph relations, low-level and high-level graph relations by layout-graph reasoning layers into a unified model, which consists of Map-to-Node module, layout-graph reasoning module and Node-to-Map module. 

\textbf{Fashion Understanding Datasets.} 
Many human-centric applications depend on reliable fashion image understanding. DeepFashion \cite{liuLQWTcvpr16DeepFashion} is a large-scale clothing dataset labeled with clothes categories, attributes, at most 8 clothes landmarks and bounding boxes. FLD~\cite{liu2016fashion} is a fashion landmark dataset (FLD) with large pose and scale variations, annotated with at most 8 landmarks and bounding boxes. Yan \etal \cite{yan2017unconstrained} contributed an unconstrained landmark dataset (ULD), which comprises 30k images with at most 8 fashion landmark annotations.
To advance the developments of domain knowledge for fine-grained fashion landmark, we propose a large-scale dataset towards the first fine-grained fashion landmark detection task, which contains 200k images annotated with at most 32 key-points for 13 clothes categories.
%-------------------------------------------------------------------------
\section{Proposed Approach}\label{sec:approach}
\subsection{Overview}
Considering the layout of fashion landmarks, we model layout-graph reasoning to enforce detected fashion landmarks be coherent with human and clothes layouts from a global perspective. We propose a model that seamlessly enforces layout relationships among landmarks on the intermediate features via multiple stacked Layout-Graph Reasoning (LGR) layers, as shown in Fig.\ref{fig:framework}.  Each LGR layer aims to map deep convolutional features into structural graph nodes via Map-to-Node module, perform reasoning over multi-level layout-graph nodes to achieve global layout coherency via a layout-graph reasoning module, and then map evolved graph nodes back to enhanced convolutional features via a Node-to-Map module. Finally a sigmoid function and a $1 \times 1$ convolution are employed to produce heatmaps of fashion landmarks. Inspired by Yang \etal \cite{yang2017learning}, we enhance intermediate features by pyramid module and decrease data bias by residual addition \cite{he2015deep}.
\subsection{Layout-graph Definition}
We define the layout-graph as a hierarchical structure for mining semantic correlations and constraints among different fashion landmarks. Specifically, we define a layout-graph constructed by graph nodes characterizing landmark categories (\eg right-collar, left-sleeve) and graph edges representing spatial layouts (\eg right-collar belongs to collar, collar and sleeve belong to upper body), which is denoted as $\mathcal{G}=\left(\mathcal{V}, \mathcal{E}\right)$. The graph nodes $\mathcal{V}$ consist of the set of leaf nodes $\mathcal{V}_{leaf}$ (\eg leaf landmark nodes) and intermediate nodes $\mathcal{V}_{middle}$ (\eg clothes-part nodes, body-part nodes, root node). We define the leaf node representations as $\mathbf{X}_{leaf}\in\mathbb{R}^{N_{leaf}\times d}$, which is generated via Map-to-Node module. We define the intermediate node representations as $\mathbf{X}_{middle}\in\mathbb{R}^{N_{middle}\times d}$, which is generated via layout-graph reasoning module. The $N_{leaf}$ and $N_{middle}$ are the number of leaf nodes and intermediate nodes, $d$ means the feature dimension of each node.

The edges $\mathcal{E}$ consist of the set of leaf edges $\mathcal{E}_{leaf}$ to represent the connections between each leaf nodes (\eg right-collar and left-collar), and intermediate edges $\mathcal{E}_{middle}$ to represent the connections between each intermediate node (\eg collar and sleeve). The leaf node adjacency weight matrix $\mathbf{A}_{leaf}\in\{0, 1\}^{N_{leaf}\times N_{leaf}}$ is initialized according to the edge connections in $\mathcal{E}_{leaf}$ as shown in Fig.\ref{fig:graph_relation}, where $0$ means disconnection and $1$ means connection. Similarly, we define the intermediate node adjacency weight matrix as $\mathbf{A}_{middle}$. In implementation, we perform normalization on all $\mathbf{A}$ by following \cite{kipf2017semi} to obtained normalized adjacency weight matrix. The $\mathbf{A}$ is crucial for layout-graph information embedded in the fashion joint layouts to benefit point-wise fashion landmark detection, which can be easily designed according to human commonsense about fashion layouts as illustrated in Sec.2 of supplementary material.
\subsection{Layout-graph Reasoning Layer}
The LGR layer aims to enhance convolutional features by layout-graph reasoning. Each LGR layer consists of three modules: 1) Map-to-Node module to map convolutional features into structural graph leaf nodes; 2) layout-graph reasoning module to model global-local clothing landmark correlations for feature enhancement, containing a graph clustering operation and a graph deconvolution operation; 3) Node-to-Map module to map evolved leaf node representations back to enhance feature representations. 
\subsubsection{Map-to-Node Module} 
This module is to seamlessly map convolutional feature maps to graph node representations. Given the input convolutional feature maps after dimension transformation ($\mathbf{F}\in\mathbb{R}^{H\times W\times C}\rightarrow \mathbf{F}\in\mathbb{R}^{HW\times C}$, where $H$, $ W$ and $C$ represent the height, weight and channel), this module produces graph leaf node representations $\mathbf{X}_{leaf}\in\mathbb{R}^{N_{leaf}\times d}$. The formulation is:
\begin{align}
\mathbf{X}_{leaf} &= \sigma(\Phi(\mathbf{FW}_m)^T\mathbf{FW}_t),\label{s3}
\end{align}
where $\mathbf{W}_m\in\mathbb{R}^{C\times N_{leaf}}$ and $\mathbf{W}_t\in\mathbb{R}^{C\times d}$ are trainable sampling matrices. The $\Phi$ denotes normalized function \textit{softmax} to sum all rows to one, and the $\sigma$ denotes non-linear function \textit{Relu}.
\subsubsection{Layout-Graph Reasoning Module}
Given graph leaf node representations $\mathbf{X}_{leaf}$ via the Map-to-Node module, our layout-graph reasoning module first performs a graph clustering operation to generate representations of intermediate nodes in the spirit of bottom-up inference, that is, propagating from (leaf landmark nodes)$\rightarrow$(clothes-part nodes)$\rightarrow$(body-part nodes)$\rightarrow$(root node). Then a graph deconvolution operation to evolve representations of bottom nodes guided by the higher-level structure nodes in the spirit of top-down inference, that is, (root node)$\rightarrow$(body-part nodes)$\rightarrow$(clothes-part nodes)$\rightarrow$(leaf landmark nodes). Benefiting from integrating the graph clustering and graph deconvolution operations, the module achieves global structural coherency.
\begin{figure*}[!t]
	\centering
	\includegraphics[width=1.0\linewidth]{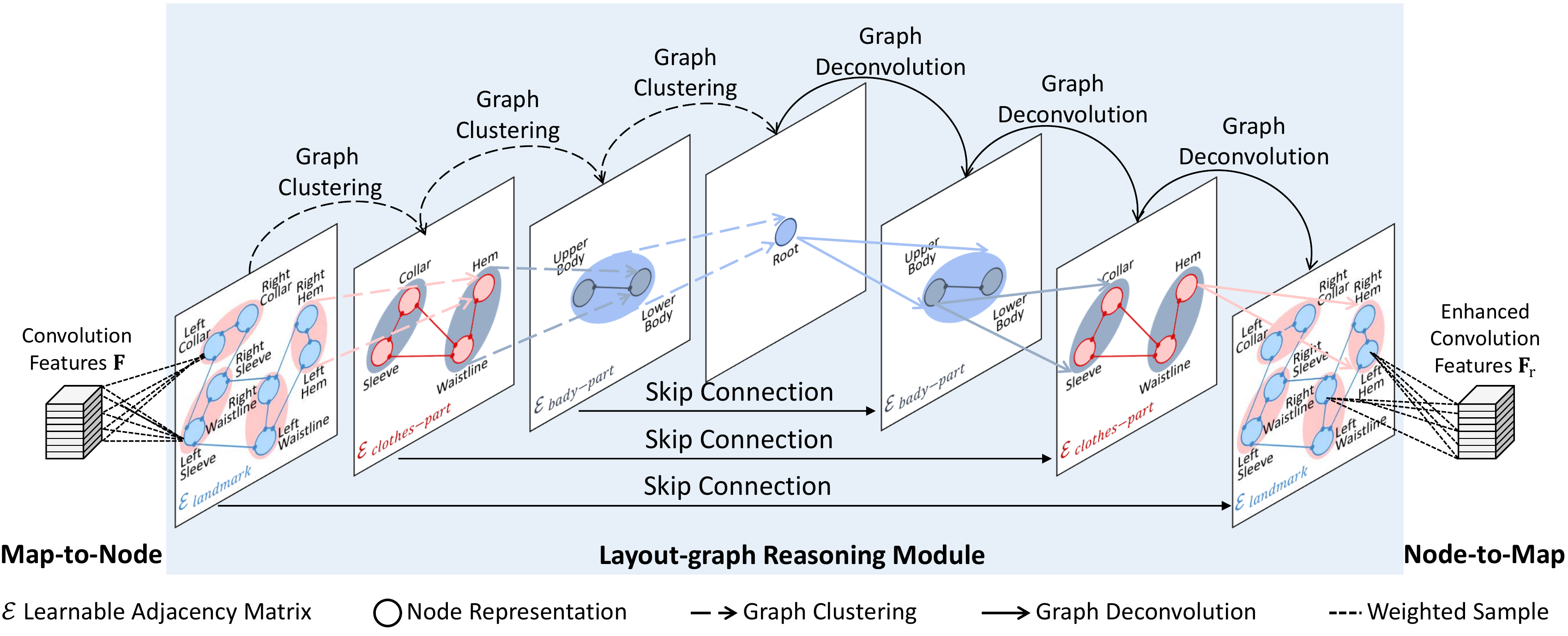}
	\vspace{-8mm}
	\caption{Illustration of our Layout-graph Reasoning (LGR) layer which contains Map-to-Node module, layout-graph reasoning module and Node-to-Map module. In Map-to-Node and Node-to-Map module, weighted sample operations vote all convolution features (evolved leaf landmark nodes) to leaf landmark nodes (enhanced convolution features) by weighted sample. In Layout-Graph Reasoning module, the graph is propagated from leaf landmark nodes to root node by Graph Clustering and Graph Reasoning. The root node is propagated back again by Graph Deconvolution and Graph Reasoning for producing evolved leaf landmark nodes. We use graph convolution from \cite{kipf2017semi} for Graph Reasoning with supervising adjacency matrix. A skip connection is employed for restricting the consistency of clustering and deconvolution operations.}
	\label{fig:reasoning}
	\vspace{-4mm}
\end{figure*}

\textbf{Graph Clustering Operation.}
This operation generates intermediate node representations by graph clustering. For different levels of bottom-up inference, the clustered graph nodes and graph edges change as shown in Fig.\ref{fig:reasoning}. The graph clustering operation of every levels is similar. Here we take $\mathbf{X}_{leaf}\rightarrow\mathbf{X}_{middle}$ as an example to illustrate the clustering operation. Given the input $\mathbf{X}_{leaf}$ and $\mathbf{A}_{leaf}$, this operation generates intermediate node representations $\mathbf{X}_{middle}$ and intermediate node adjacency weight matrix $\mathbf{A}_{middle}$, which is formulated as:
\begin{align}
\mathbf{X}_{middle} &= \sigma\left(\mathbf{W}_{p'}^T\mathbf{A}_{leaf}\mathbf{X}_{leaf}\mathbf{W}_h\right) \label{gsp1},\\
\mathbf{A}_{middle} &= \sigma\left(\mathbf{W}_{p}^T\mathbf{A}_{leaf}\mathbf{W}_p\right)\label{gsp2},
\end{align}
where $\mathbf{W}_{p'}\in\mathbb{R}^{N_{leaf}\times N_{middle}}$ and $\mathbf{W}_p\in\mathbb{R}^{N_{leaf}\times N_{middle}}$ are both trainable clustering matrices. The $\mathbf{W}_h\in\mathbb{R}^{d\times d}$ is a trainable weight matrix. We use graph convolution from \cite{kipf2017semi} for graph reasoning to perform over $\mathbf{X}_{leaf}$ and $\mathbf{A}_{leaf}$ using $\mathbf{W}_h$ to update the leaf graph node representations. Then we utilize $\mathbf{W}_{p'}$ to cluster updated $\mathbf{X}_{leaf}$ into $\mathbf{X}_{middle}$, which is formulated as Eq.\ref{gsp1}. We cluster $\mathbf{A}_{leaf}$ using $\mathbf{W}_{p}$ to generate $\mathbf{A}_{middle}$, which is formulated as Eq.\ref{gsp2}. $\mathbf{W}_p$ is a permutation matrix and obeys distribution of $\mathbf{W}_p^T\mathbf{W}_p=\mathbf{I}$. We use graph reasoning to perform over $\mathbf{X}_{middle}$ and $\mathbf{A}_{middle}$ to update the clustered graph node representations $\mathbf{X}_{middle}\in\mathbb{R}^{N_{middle}\times d}$. 

\textbf{Graph Deconvolution Operation.} This operation evolves representations of bottom nodes guided by the higher-level structure nodes in the spirit of top-down inference as shown in Fig.\ref{fig:reasoning}. Again, we take $\mathbf{X}_{middle}\rightarrow\mathbf{X}_{leaf}$ an as example to illustrate the deconvolution operation. Given the input intermediate node representations $\mathbf{X}_{middle}$ and adjacency matrix $\mathbf{A}_{middle}$ from higher-level structure, we utilize the formulation like Eq.\ref{gsp1} and Eq.\ref{gsp2} to produce leaf node representations $\mathbf{X}_{leaf}$ and leaf node adjacency weight matrix $\mathbf{A}_{leaf}$. Furthermore, to integrate the high-level and low-level structure information, we utilize a matrix addition over the node representations before clustering and after deconvolution, followed by the graph reasoning to update the evolved leaf node representations $\mathbf{X}_{leaf}$.
\subsubsection{Node-to-Map Module}\label{node-to-map}
We map evolved graph nodes into enhanced convolutional features via Node-to-Map module. Given the input convolutional features $\mathbf{F}$ and evolved leaf node representations $\mathbf{X}_{leaf}$, this module aims to generate enhanced convolutional feature representations $\mathbf{F}_r$. We first perform the dimension transformation for $\mathbf{F}\in\mathbb{R}^{HW\times C}\rightarrow\mathbf{F}\in\mathbb{R}^{H W\times N\times C}$ and $\mathbf{X}_{leaf}\in\mathbb{R}^{N_{leaf}\times d}\rightarrow\mathbf{X}_{leaf}\in\mathbb{R}^{HW\times N_{leaf}\times d}$. Then we concatenate $\mathbf{F}$ and $\mathbf{X}_{leaf}$ to  $\mathbf{X}_a\in\mathbb{R}^{HW\times N_{leaf}\times (C+d)}$ for richer feature representations. We formulate this module as:
\begin{align}
\mathbf{F}_r &= \sigma(\Phi(\mathbf{X}_a\mathbf{W}_{m'})\sigma(\mathbf{X}_{leaf}\mathbf{W}_{t'})) \label{nm},
\end{align}

Eq~\ref{nm} is to map node representations $\mathbf{X}_{leaf}\in\mathbb{R}^{N_{leaf}\times d} $ to enhanced convolutional features $\mathbf{F}_r\in\mathbb{R}^{HW\times C}$, where $\mathbf{W}_{m'}\in\mathbb{R}^{C+d}$ is a vector with $C+d$ dimension and $\mathbf{W}_{t'}\in\mathbb{R}^{d\times C}$ is a trainable sampling matrices.
\section{Experiments}
\subsection{Experimental Settings}
\textbf{Network Architecture.} 
Following the baseline of \cite{liu2016deepfashion,liu2016fashion,yan2017unconstrained,wang2018attentive}, we use VGG16 \cite{simonyan2014very} with four stacked LGR layers for feature extraction and layout-graph reasoning. Each LGR layer contains Map-to-Node module, layout-graph reasoning module and Node-to-Map module. We map convolutional features into graph leaf node representations via Map-to-Node module. On DeepFashion and FLD, we set 8 leaf nodes (\eg left-collar, right-hem), 6 intermediate nodes including 4 clothes-part nodes (collar, hem) and 2 body-part nodes (\eg upper body, lower body), and 1 root node. On FFLD, we set 32 leaf nodes (\eg left-shoulder, crotch), 14 intermediate nodes including 12 clothes-part nodes (\eg sleeve, knee) and 2 body-part nodes (\eg upper body, lower body), and 1 root node. More details of node's layout can be seen in supplementary material. Then we model layout-graph of fashion landmarks via layout-graph reasoning module to evolve graph node representations guiding by defined graph correlations as shown in Fig.\ref{fig:graph_relation}. The Node-to-Map module to map evolved graph node representations into convolutional features for enhancing the feature representations, which results are fed into a 1$\times$ 1 convolution with $sigmoid$ activation to get the prediction. A residual addition and pyramid feature post-processing are appended between each LGR layer for reducing bias and capturing rich representations across multi-scales.
\begin{table*}[t]
	\centering
	%		\normalsize
	\footnotesize
	%		\scriptsize
	\caption{Comparison with the state-of-the-art model on the FLD test set and DeepFashion test set using the NE metric.}
	\vspace{-2mm}
	\label{tab: fld_test_landmark}
	\begin{tabular}{c|ccccccccc}
		\hline
		\multicolumn{10}{c}{FLD} \\ \hline
		Methods                                      & L.Collar  & R.Collar & L.Sleeve & R.Sleeve & L.Waistline  & R.Waistline  & L.Hem & R.Hem & Avg.  \\ \hline 
		FashionNet \cite{liu2016deepfashion}           & .0784   & .0803     & .0975  & .0923 & .0874 & .0821 & .0802  & .0893   & .0859\\
		PyraNet \cite{yang2017learning}  &.0341 &.0341&.0610 &.0620 &.0920 &.0921 &.0314 &.0291  &.0723\\
		DFA \cite{liu2016fashion}                    & .048  & .048     & .091  & .089  & - & - & .071  & .072   & .068 \\
		DLAN \cite{yan2017unconstrained}          & .0531  & .0547     & .0705  & .0735  & .0752 & .0748 & .0693  & .0675   & .0672\\
		BCRNNs \cite{wang2018attentive}          & .0463  & .0471     & .0627  & \textbf{.0614}  & .0635 & .0692 & .0635  & .0527  & .0583 \\  \hline
		LGR(ours)                        & \textbf{.0423}  & \textbf{.0152} & \textbf{.0502} & .0735 & \textbf{.0195} & \textbf{.0512} & \textbf{.0452} & \textbf{.0393} &\textbf{.0419}\\ \hline \hline
		\multicolumn{10}{c}{DeepFashion} \\ \hline
		Methods            & L.Collar  & R.Collar & L.Sleeve & R.Sleeve & L.Waistline  & R.Waistline  & L.Hem & R.Hem & Avg.  \\ \hline 
		FashionNet \cite{liu2016deepfashion}           & .0854  & .0902     & .0973  & .0935  & .0854 & .0845 & .0812  & .0823   &.0872 \\
		PyraNet \cite{yang2017learning}  &.0343 &.0343 &.0602 &.0613 &.0920 &.0931 &.0308 &.0291  &.0719\\
		DFA \cite{liu2016fashion}                     & .0628  & .0637     & .0658  & .0621  & .0726 & .0702 & .0658  & .0663   & .0660 \\
		DLAN \cite{yan2017unconstrained}           & .0570  & .0611     & .0672  & .0647  & .0703 & .0694 & .0624  & .0627   &.0643\\
		BCRNNs \cite{wang2018attentive}          & .0415  & .0404     & .0496  & .0449  & .0502 & .0523 & .0537  & .0551   & .0484\\  \hline
		LGR(ours)            & \textbf{.0270}  & \textbf{.0116} & \textbf{.0286} & \textbf{.0347} & \textbf{.0307} & \textbf{.0435} & \textbf{.0160} & \textbf{.0162} &\textbf{.0336}\\
		\hline
	\end{tabular}
%	\vspace{-6mm}
\end{table*}

\textbf{Three Benchmarks and Evaluation.} 
We evaluate and report the results and comparisons on three datasets. DeepFashion \cite{liu2016deepfashion} is the largest fashion dataset so far, which contains 289,222 images annotated with bounding boxes and at most 8 landmarks. FLD \cite{liu2016fashion} is a fashion landmark dataset with more diverse variations (\eg pose, scale, background), which contains 123, 016 images annotated at most 8 landmarks and bounding boxes per image. FFLD is our contributed fine-grained fashion landmark dataset, which contains 200k images annotated with at most 32 key-points and bounding boxes for 13 clothes categories. Following \cite{wang2018attentive}, 209,222 fashion images are used for training; 40, 000 images are used for validation and remaining 40, 000 images are for testing in DeepFashion. Following the protocol in FLD \cite{liu2016fashion}, 83, 033 images and 19, 992 fashion images are used for training and validating, 19, 991 images are used for testing. In FFLD, we use 120K images as a training set, 40K images as a validation set and 40K images as a test set. Normalized error(NE) metric \cite{liu2016fashion} is adopted for evaluation. We utilize $l_2$ function to calculate the distance between predicted heatmaps and ground-truth in normalized coordinate space (\ie normalized by the height and width of image).
\begin{figure*}[t]
	\centering\includegraphics[width=0.9\linewidth]{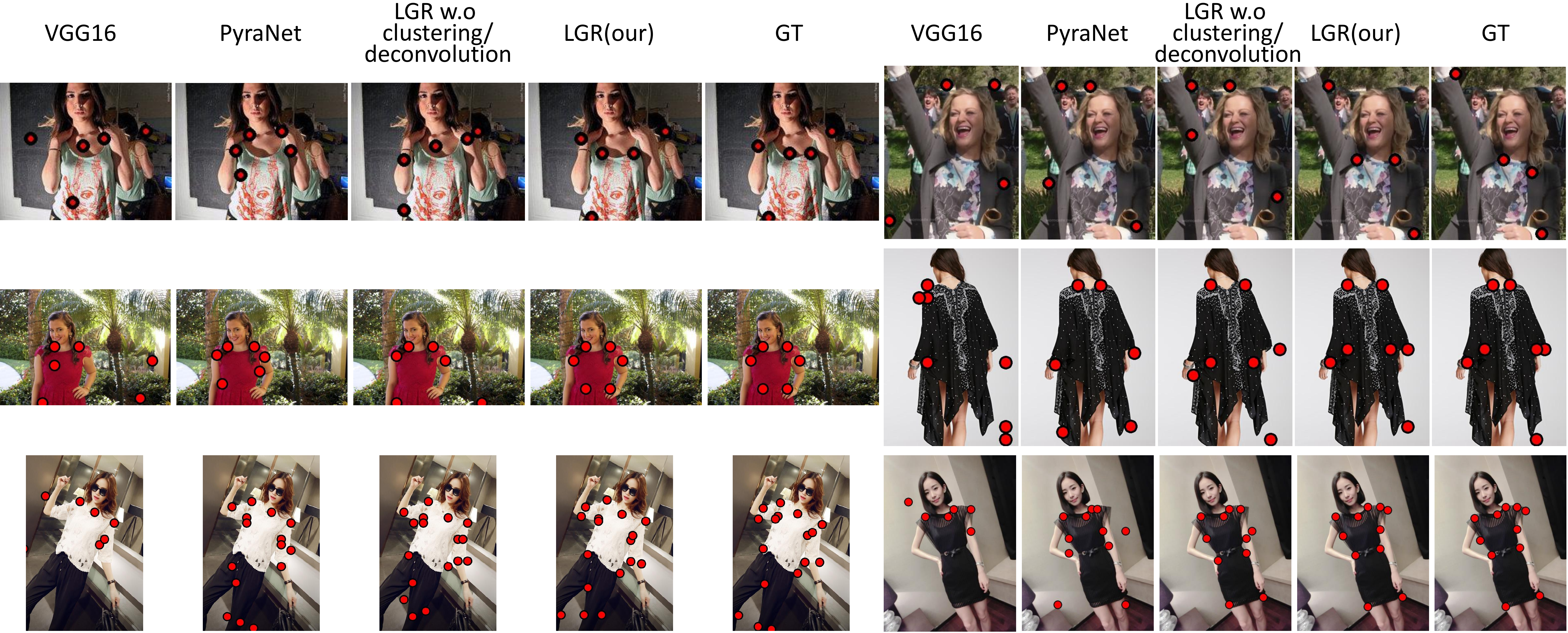}
	\vspace{-3mm}
	\caption{Qualitative results for VGG16 \cite{simonyan2014very}, PyraNet \cite{yang2017learning}, LGR w.o clustering/deconvolution (two graph layers without graph clustering and deconvolution) and LGR over DeepFashion (first row), FLD (second row) and FFLD (bottom row). The detected landmarks (red circle) are performed on different variations such as occlusion and complicate background. Please see the zoomed-in color pdf file.}
	\label{fig:result2}
	\vspace{-6mm}
\end{figure*}

\textbf{Training Strategy and Object Function.} 
We use LGR layer for fashion landmark detection over FLD \cite{liu2016fashion}, DeepFashion \cite{liu2016deepfashion} and FFLD separately without any pre-trained models. Following \cite{wang2018attentive}, we first crop each image using labeled bounding boxes, resize the cropped image to 224 $\times$ 224, and extract features for graph reasoning. The training data are augmented by scaling, rotation, and flipping. We train all the models using stochastic gradient descent with a batch size of 16 images, which is optimized by Adam optimizer \cite{kingma2014adam} with an initial learning rate of 1.e-3 on an 11 GB NVIDIA 1080Ti GPU. Betas of Adam are from 0.9 to 0.999. On FLD, we linearly drop the learning rate by a factor of 10 every 20 epochs. On DeepFashion and FFLD, we linearly decrease the learning rate by a factor of 10 every 10 epochs. We stop training when no improvements on the validation set. We set the mean squared error (MSE) equation as an objective function between the final predicted heatmaps and ground-truth.
\subsection{Comparison with the state-of-the-arts}
LGR achieves an obvious improvement over the two large datasets compared with PyraNet \cite{yang2017learning}, FashionNet \cite{liu2016deepfashion}, DFA \cite{liu2016fashion}, DLAN \cite{yan2017unconstrained} and BCRNNs \cite{wang2018attentive}. Note that PyraNet is human pose estimation model with two stages. We train the PyraNet following the same strategy as \cite{yang2017learning}. Our LGR outperforms SOTA at 0.0419 on FLD and 0.0336 on DeepFashion, which is much lower than the closest competitor (0.0583 and 0.0484), as shown in Table.\ref{tab: fld_test_landmark}. Compared with traditional DCNNs \cite{liu2016deepfashion} and grammar model \cite{wang2018attentive}, we further model layout-graph reasoning to enforce detected fashion landmarks be coherent with human and clothes layouts from a global perspective. Benefiting from the joint reasoning with hierarchical structures of fashion landmarks, we achieve the SOTA performs over all existing  models by a large improvement. Note that our model consistently decreases the NE in all landmarks on DeepFashion.
\subsection{Ablation Study}
\begin{table}[t]
	\centering
	%		\normalsize
	\footnotesize
	\caption{Ablation study on FLD and DeepFashion using NE metric (Avg.). The structures with different numbers of graph clustering and deconvolution are shown in Fig.\ref{fig:pooling}. We also present the results generate by different numbers of normal graph convolutional layers that replacing the graph clustering and deconvolution. We also compare the average execution time for testing (Time).
	}
	\label{tab: ablation2}
	\vspace{-2mm}
	\begin{tabular}{c|cc|cc}
		\hline
		\multicolumn{5}{c}{\textbf{Different Stack numbers}}\\\hline
		&\multicolumn{4}{c}{FLD}\\\hline
		Methods &Avg.& $\Delta$Avg. & Time(s) &$\Delta$Time(s) \\ \hline
		VGG16 \cite{simonyan2014very} & .0871 & .0452&\textbf{.00065} &.00314 \\ 
		one stack &.0711 &.0292 &.00155 &.00224 \\ 
		two stacks & .0535 &.0116  & .00236 &.00143\\ 
		three stacks &.0529 &.0110  &.00346 &.00033\\ 
		\textbf{four stacks} &.0419 &-  &.00379 &-\\ 
		five stacks &\textbf{.0405} &.0014  &.00472 &.00093\\
		\hline\hline
		\multicolumn{5}{c}{\textbf{Different Graph Layers}} \\\hline
		& \multicolumn{2}{c|}{FLD} & \multicolumn{2}{c}{DeepFashion}\\ \hline
		Methods & Avg. &Time(s) &Avg. &Time(s) \\\hline
		one-layer &.0531 &\textbf{.00241} &.0482 &\textbf{.00237}\\
		two-layer &\textbf{.0471} &.00273 &\textbf{.0437} &.00266\\
		four-layer &.0639 &.00279 &.0562 &.00271\\
		six-layer &.0644 &.00289 &.0638 &.00300\\
		eight-layer &.0954 &.00357 &.0779 &.00341\\
		\hline\hline
		\multicolumn{5}{c}{\textbf{Different Number of Graph Clustering and Deconvolution}} \\\hline
		& \multicolumn{2}{c|}{FLD} & \multicolumn{2}{c}{DeepFashion}\\ \hline
		Methods & Avg. &Time(s) &Avg. &Time(s) \\\hline
		one-clustering &.0488 &\textbf{.00267} &.0403 &\textbf{.00261}\\
		two-clustering &.0443 &.00302 &.0372 &.00336\\
		\textbf{three-clustering} &\textbf{.0419} &.00379 &\textbf{.0336} &.00352\\
		\hline\hline
		\multicolumn{5}{c}{\textbf{Different Injected Layers}}\\\hline
		&\multicolumn{4}{c}{FLD}\\\hline
		Methods & \multicolumn{4}{c}{Avg.} \\\hline
		VGG16 ConvBlock1 &\multicolumn{4}{c}{.0811}\\
		VGG16 ConvBlock3 &\multicolumn{4}{c}{.0574}\\
		VGG16 ConvBlock5 &\multicolumn{4}{c}{\textbf{.0419}}\\\hline
	\end{tabular}
	\vspace{-6mm}
\end{table}
\textbf{Different Stack Numbers.}
There are six experiments to display the performance of different stacked LGR layers, which are shown in the first list of Table.\ref{tab: ablation2}. VGG16 without any graph reasoning achieves 0.0871 average NE, which is the worst result compared with other knowledge-guide model. Comparing the variants of different stages, the performances get better with the stack increasing, which is a coarse-to-fine processing to repeatedly refining the prediction. Five stacked LGR gets the best performance(0.0405 NE), while it needs more GPU memory and time during training/validating/testing progress. The gap of all landmarks between five stacks and four stacks are closed. Limited by device and time, we select four stacks in our stand model and apply it to all extensive experiments.

\textbf{Different Graph Layers.}
In the second list of Table.\ref{tab: ablation2}, we built an ablation study of different normal graph layers on FLD and DeepFashion. To demonstrate the superior ability of graph clustering and graph deconvolution, we replace a LGR layer with one graph layer without graph clustering and deconvolution, which is presented as one-layer. Benefiting from graph reasoning, two graph layers can get the best performance, while the performance tends to be destroyed with the depth increasing of graph layers (\eg two-layer:0.0471, eight-layer:0.0954). Purely increasing the graph layers can not simply get better performance, but the spent time also grows up with the model size increasing. LGR can attenuate the shortcoming as above and get better performance with layer increasing by graph clustering and graph deconvolution. Compared the LGR layer with normal graph layer, eight graph layers got worse performance compared with three-clustering (0.0954 and 0.0419). Note that three-clustering got close speed compared with eight graph layers (\eg 0.00379s and 0.00357s). Due to the LGR layer contains graph clustering, graph deconvolution and graph reasoning operations, which has more processes compared with the same size of normal graph layers.

\begin{figure}[t]
	\centering\includegraphics[width=1.0\linewidth]{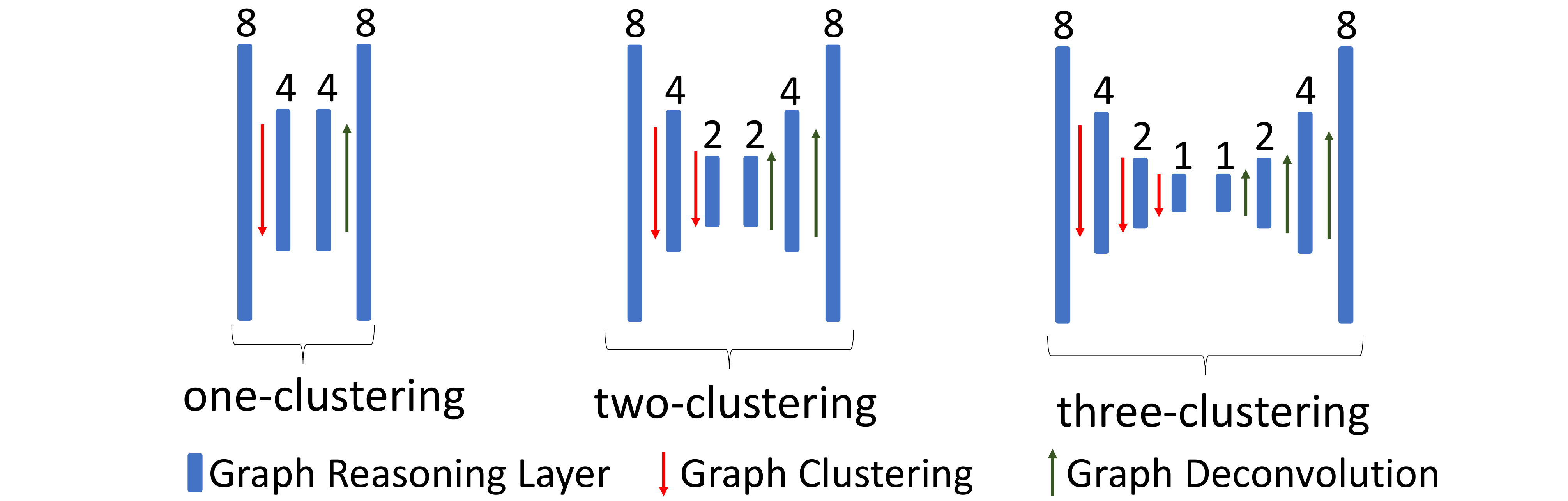}
	\vspace{-3mm}
	\caption{The different structures of graph clustering and graph deconvolution. The number of nodes in each graph reasoning layer is labeled on the top. Please see the zoomed-in color pdf file.}
	\label{fig:pooling}
	\vspace{-6mm}
\end{figure}

\textbf{Different Number of Graph Clustering and Deconvolution.}
In the third list of Table.\ref{tab: ablation2}, we explore the effectiveness of different numbers of graph clustering operation and graph deconvolution operation in LGR layer, the diagram as shown in Fig.\ref{fig:pooling}. With the depth of operations increasing, more prior commonsense knowledge in terms of richer human body part layouts and clothes part layouts will be got to better guide the learning processing. The experiment results have shown that three-clustering get better performance than one clustering (0.0419 and 0.0488 on FLD, 0.0336 and 0.0403 on DeepFashion).

\textbf{Different Injected Layers.}
The fourth list of Table.\ref{tab: ablation2} compares the variants of injecting four stacked LGR layers into different convolution blocks (ConvBlock) of VGG16 \cite{simonyan2014very} over FLD. The stacked LGR layers are injected into right before the block. The performance of adding LGR layers after Block3 is worse than adding LGR layers after ConvBlock5 of VGG16. We show the possible reason that the deeper layers can encode more semantically high-level feature representation, which is more suitable for layout-graph reasoning.

\subsection{Qualitative Results}
The results showed different abilities of traditional DCNNs \cite{simonyan2014very}, PyraNet \cite{yang2017learning}, normal graph reasoning without graph clustering and deconvolution and LGR. We select the best structure (two-layer) of normal graph layers demonstrated as above. In Fig.\ref{fig:result2}, for complex background, diverse clothes layouts and styles, multiple scales and views, the pure DCNNs, pose estimation model \cite{yang2017learning} and normal graph layers fail to detect correct fashion landmark. Benefiting from modeling layout-graph relations of landmarks by a hierarchical structure, LGR can mine semantic coherency of layout-graph and enhance the semantic correlations and constrains among landmarks (\eg collar and sleeve belong to upper body). LGR covers difficult variance and generate reasonable results guided by layout-graph reasoning during the bottom-up, top-down inference. For example, LGR can detect correct results by constraining fashion landmarks on one clothing in complex background (first row in Fig.\ref{fig:result2}). 
\subsection{Fine-grained Fashion Landmark Dataset (FFLD)} 
Compared with exited fashion landmark datasets \cite{liu2016deepfashion,liu2016fashion,yan2017unconstrained}, FFLD consists of more than 70\% consumer images, which is more challenge for multiple view and light, complex background and deformable clothes appearance. Note that the FFLD is the closest fashion landmark dataset with the real application. More detailed definition and statistics of FFLD can be seen in supplementary material.

We have shown four existing methods evaluated on FFLD in Table \ref{tab: ffld} to comprehensively perform FFLD. The FashionNet \cite{liu2016deepfashion} and BCRNN \cite{wang2018attentive} are the SOTA methods for fashion landmark detection, and the PyraNet \cite{yang2017learning} is one of SOTA methods for human pose estimation. We utilize VGG16 \cite{simonyan2014very} stacked with two graph convolutional layers, which is regard (GCN) \cite{kipf2017semi} is a typical graph-based methods, which VGG16 in this evaluation. Based on the prior layout-graph as shown in Fig.\ref{fig:graph_relation}, layout-graph reasoning with four stacks is evaluated on FFLD. 

As shown in Table.\ref{tab: ffld}, the LGR achieved 0.118 average NE on FFLD, which is a worse performance compared with FLD(0.0419 NE) and DeepFashion(0.0336 NE) due to more consumer images, more fine-grained fashion landmarks, more challenge views and backgrounds. To demonstrate the challenge of FFLD on other models, we perform fashion landmark detection model (FashionNet \cite{liu2016deepfashion} and BCRNN \cite{wang2018attentive}), human pose estimation model (PyraNet \cite{yang2017learning}) and normal graph layer (GCN \cite{kipf2017semi}) on FFLD to achieve 0.2031 NE, 0.1226 NE, 0.1423 NE and 0.1272 NE. We perform BCRNN on FFLD following the setting of \cite{wang2018attentive}, and the fashion landmark grammars of FFLD consist of kinematics grammar and symmetry grammar. More detailed fashion grammars of FFLD can be seen in supplementary material.
\begin{table}[t]
	\centering
	\tabcolsep 0.05in{\scriptsize{}}
	%			\normalsize
	\caption{Evaluation of different models on FFLD.}
	\vspace{-2mm}
	\label{tab: ffld}
	\begin{tabular}{c|c}
		\hline
		Methods & Avg. NE\\ \hline
		FashionNet \cite{liu2016deepfashion}   & .2031 \\
		PyraNet \cite{yang2017learning} & .1423 \\
		GCN \cite{kipf2017semi} & .1272 \\
		BCRNN \cite{wang2018attentive} &  .1226\\
		LGR (ours) & \textbf{.1180}\\ 
		\hline
	\end{tabular}
	\vspace{-6mm}
\end{table}
\section{Conclusion}
In this paper, we proposed the Layout-Graph Reasoning (LGR) that consists of three modules for fashion landmark detection to seamlessly utilize structural graph reasoning in a hierarchical way. We use LGR to achieve SOTA performance over recent methods. We contribute a fine-grained fashion landmark dataset to advance the development of knowledge graph to fashion landmark research.
\section{Acknowledgements}
This work was supported in part by the Sun Yat-sen University Start-up Foundation Under Grant No. 76160-18841201, in part by the National Key Research and Development Program of China under Grant No. 2018YFC0830103, in part by National High Level Talents Special Support Plan (Ten Thousand Talents Program), and in part by National Natural Science Foundation of China (NSFC) under Grant No. 61622214, and 61836012.

{\small
\bibliographystyle{ieee}
\bibliography{egbib}
}

\end{document}